\documentclass[conference]{IEEEtran}
\IEEEoverridecommandlockouts

\usepackage{algorithmic}
\usepackage{amsmath, amssymb, amsfonts}
\usepackage[english]{babel} 
\usepackage{cite}
\usepackage[hidelinks]{hyperref}
\usepackage{textcomp}
\usepackage[small, labelsep=period]{caption}
\usepackage[pdftex]{graphicx}
\usepackage[hidelinks]{hyperref} 
\usepackage{paralist}
\usepackage{quotmark}
\usepackage[usenames, dvipsnames, table]{xcolor}

\usepackage{soul}
\usepackage{nicematrix}
\usepackage{tikz}
\usetikzlibrary{shapes.geometric, shapes.misc, arrows, positioning}
\usepackage{pgfplots}
\usepgfplotslibrary{fillbetween}


\usepackage{xpatch}
\xpatchcmd\IEEEkeywords{---}{-}{}{}

\makeatletter
\renewcommand{\fnum@figure}{Figure~\thefigure}
\makeatother

\newcommand{\ie}{i.e., }

\newcommand{\domainX}{1}
\newcommand{\domainY}{\domainX * 2.6}

\colorlet{TrustorColor}{blue}
\colorlet{TrusteeColor}{Orange}
\colorlet{CategoryColorReliability}{blue!80}
\colorlet{CategoryColorSafety}{red!50}
\colorlet{CategoryColorSecurity}{brown}
\colorlet{CategoryColorAccountability}{violet!90}
\colorlet{CategoryColorExplainability}{orange!70}
\colorlet{CategoryColorPrivacy}{cyan!50}
\colorlet{CategoryColorFairness}{teal}

\def\BibTeX{{\rm B\kern-.05em{\sc i\kern-.025em b}\kern-.08em
    T\kern-.1667em\lower.7ex\hbox{E}\kern-.125emX}}
    
\begin{document}

\setlength{\abovedisplayskip}{-5pt}
\setlength{\belowdisplayskip}{3pt}

\title{\bfseries\Large Fostering Trust and Quantifying Value of AI and ML}

\author{
\IEEEauthorblockN{Dalmo Cirne}
\IEEEauthorblockA{\textit{Machine Learning for Financials} \\
    \textit{Workday}\\
    Boulder, Colorado, USA \\
    email: dalmo.cirne@workday.com}
\and
\IEEEauthorblockN{Veena Calambur}
\IEEEauthorblockA{\textit{Responsible AI} \\
    \textit{Workday}\\
    Princeton, New Jersey, USA \\
    veena.calambur@workday.com}
}

\maketitle

\begin{abstract}
Artificial Intelligence (AI) and Machine Learning (ML) providers have a responsibility to develop valid and reliable systems. Much has been discussed about trusting AI and ML inferences (the process of running live data through a trained AI model to make a prediction or solve a task), but little has been done to define what that means. Those in the space of ML-based products are familiar with topics such as transparency, explainability, safety, bias, and so forth. Yet, there are no frameworks to quantify and measure those. Producing ever more trustworthy machine learning inferences is a path to increase the value of products (\ie increased trust in the results) and to engage in conversations with users to gather feedback to improve products. In this paper, we begin by examining the dynamic of trust between a provider (Trustor) and users (Trustees). Trustors are required to be trusting and trustworthy, whereas trustees need not be trusting nor trustworthy. The challenge for trustors is to provide results that are good enough to make a trustee increase their level of trust above a minimum threshold for: 1- doing business together; 2- continuation of service. We conclude by defining and proposing a framework, and a set of viable metrics, to be used for computing a \emph{trust score} and objectively understand how trustworthy a machine learning system can claim to be, plus their behavior over time.
\end{abstract}

\begin{IEEEkeywords}
\textbf{\textit{artificial intelligence, machine learning, trust, game theory.}}
\end{IEEEkeywords}

\section{Introduction}
\label{section:introduction}

Much has been said about responsible Artificial Intelligence (AI), but the majority of those conversations are high-level and focused on defining principles---which are important for defining direction---but are rarely coupled with the actual operation of ML-based systems.

Measuring the increase or decrease of trust in this technology is a gap that needs to be addressed, and that is the main proposal of this paper: a quantitative framework to be used in computing the trustworthiness of AI and ML systems. Here, trust is defined as the willingness to interact with an AI/ML system while being aware that a model inference \cite{what-is-ai-inference} is fallible.

The framework, however, is not without its challenges. There are several other elements to be considered in an AI/ML-powered system in order for it to gain the trust of its users. Good inferences are one of them, but so is data privacy, mitigating bias, measuring qualitative aspects, tracking the trust level over time, model training automation, and so on.

The paradigm explored in this paper assumes that trust is built by the trustor's initial act, signaling that the actor is trustworthy. More specifically, the trustor's act would be to invest in building a product and offer it to customers with the promise that it will generate value to them; more value than what is paid in return for the service. The trustor decides how much to invest, and the trustee decides whether to reciprocate and give continuity to the business relationship.

Note that the trustee does not have to be held to similar standards for trustworthiness as the trustor. The objective is to make the customers trusting---above a minimum threshold $T$---as to engage in the \emph{Trust Games}\cite{BERG1995122}. These games are extensions built on top of the \emph{Game Theory}\cite{game-theory}. Furthermore, trust has a temporal element to it. Once established, there are no guarantees that there will be a continuation. Therefore, this is an extensive form of interaction where both actors collaborate and observe each other, reacting to historical actions from one another.

A global study, conducted by the services and consulting firm KPMG, and named “\href{https://assets.kpmg.com/content/dam/kpmg/au/pdf/2023/trust-in-ai-global-insights-2023.pdf}{Trust in Artificial Intelligence}\cite{trust-in-AI},” has found that there is a wariness sentiment in large sections of the workforce in general. The people surveyed in the study expressed concern about trusting those systems, from financials to human capital management products. The framework proposed in this paper will help address such sentiment by quantifying and measuring trust in AI and ML. The results can then be shared with the workforce or the population as a whole to help them better understand how ML-based solutions function and in turn, develop a positive sentiment towards adopting such products.

The rest of the paper is structured as follows. In Section II, we examine the dynamic of trust between a provider (Trustor) and users (Trustees). In Section III, we propose a quantification of trust over many iterations between trustor and trustee. In Section IV, we define a minimum trust threshold. In Section V, we present simulations of the quantification of trust. In Section VI, we present the categories for measuring trust. In Section VII, we demonstrate how the trust score can be practically implemented. In Section VIII, we define a region of fair trading between trustor and trustee. Section IX concludes our work.

\section{Trust Games}
\label{section:trust_games}

The motion of a \emph{trust game} is developed around two actors: a trustor and a trustee. The trustor has a service of value $V$ to offer to a trustee. The value in question is \emph{quality machine learning inferences}. ML is implemented as a software service, and by its nature, software can be replicated to any number $n$ of customers without physical constraints. Thus, $V$ can be offered independently and concurrently to all customers.

It could be the case that the value $V$ of inferences may be only partially absorbed by a trustee. The limited, portioned consumption could be due to a variety of reasons, including, but not limited to: \begin{inparaitem}[]
    \item eligibility or capacity to use all the features (\ie satisfies all requirements),
    \item service subscription tiers,
    \item users have yet to be trained.
\end{inparaitem}

In order to represent the range of scenarios where the trustor may transfer the entirety of value $V$ or a smaller portion of it, we introduce a multiplier $p$, where $\{p \in \mathbb{R} \mid 0 \leq p \leq 1 \}$. Therefore, the initial remittance sent by trustor $u$ is:

\begin{align} \label{eq:trustor_initial_remittance}
	R_u = p V
\end{align}

Depending on the quality of the trustor's results, trustees' perception of value may be magnified or reduced by a factor $K$, where $\{ K \in \mathbb{R} \}$. For $K > 1$, it means that the trustor improved the efficiency of operations for the trustee (they do better than operating on their own). For $K = 1$, the trustee is operating at the same efficiency, and for $K < 1$ (negative values are also possible) the trustee is less efficient than before they started using the service. The initial perceived gain received by trustee $v$ is:

\begin{align} \label{eq:trustee_initial_receival}
	G_v &= K R_u \notag \\
	    &= K p V
\end{align}

A trustee is free to reciprocate or not. During a trial period, they may choose to decline further service. Even if under contract, they may choose to skip renewal. On the other hand, assuming that the value received from ML inferences improved their efficiency, the incentive is to continue to engage. In either case, a trustee will give back a portion $q$ of the gain received, where $\{q \in \mathbb{R} \mid 0 \leq q < 1 \}$. The value sent back may take the form of monetary payment for the service, interviews, usability feedback, labeling of transactions, or a combination of those. The repayment $B$ expected by trustor $u$ is therefore:

\begin{align} \label{eq:repayment_trustee}
	B_u &= q G_v \notag \\
	    &= q K p V
\end{align}

There could be a consideration to introduce a magnification factor on the repayment from trustee $v$. That, however, is not necessary in the scope of this paper since trustees do not need to be trustworthy; the trustor $u$ is not evaluating whether to trust them or not.

Fig.~\ref{fig:trust_game_payoffs} represents the flow of the initial step in this trust game. The {\color{TrustorColor} blue line} segment represents the range of possible values delivered to trustees by the trustor, the large {\color{TrustorColor} blue circle} is the magnification factor applied to the value delivered, and the {\color{TrusteeColor} orange line} segment represents the range of possible values reciprocated to the trustor by a trustee.

{
\tikzstyle{circle_node}=[
    circle,
    minimum size=2pt,
    draw=black,
    fill=black
]

\tikzstyle{circle_trustor}=[
    circle,
    minimum size=1pt,
    draw=TrustorColor,
    fill=TrustorColor
]

\tikzstyle{circle_trustee}=[
    circle,
    minimum size=1pt,
    draw=TrusteeColor,
    fill=TrusteeColor
]

\begin{center}
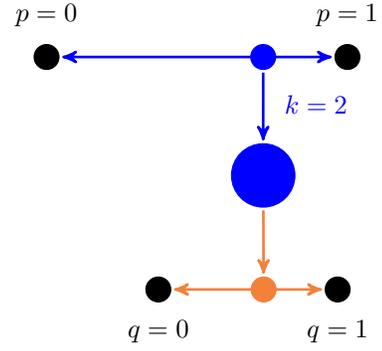

\begin{tikzpicture}[-, >=stealth', shorten <= 1pt, shorten >= 1pt, auto, node distance=14mm, minimum width=12mm,
                    minimum height=6mm, align=center, scale=0.8]
    
    \node[circle_node] (trustor_min_val) at (-2.5, 0) {};
    \node[above of=trustor_min_val, yshift=-24pt] {$p=0$};
    \node[circle_node] (trustor_max_val) at (2.5, 0) {};
    \draw[<->, line width=1pt, color=TrustorColor] (trustor_min_val) to (trustor_max_val);
    \node[above of=trustor_max_val, yshift=-24pt] {$p=1$};
    \node[circle_trustor] (trustor_value) at (1.1, 0) {};
    \node[below of=trustor_value, color=TrustorColor, xshift=20pt, yshift=22pt] {$k=2$};
    
    \node[circle_trustor, below of=trustor_value, minimum size=24pt, yshift=-5pt] (magnification) {};
    \draw[->, line width=1pt, color=TrustorColor] (trustor_value) to (magnification);
    
    \node[circle_node, below left of=magnification, xshift=-11.5pt, yshift=-15pt] (trustee_f_0) {};
    \node[below of=trustee_f_0, yshift=24pt] {$q=0$};
    \node[circle_node, right of=trustee_f_0, xshift=28pt] (trustee_f_1) {};
    \node[below of=trustee_f_1, yshift=24pt] {$q=1$};
    \draw[<->, line width=1pt, color=TrusteeColor] (trustee_f_0) to (trustee_f_1);
    \node[circle_trustee, right of=trustee_f_0] (trustee_feedback) {};
    \draw[->, line width=1pt, color=TrusteeColor] (magnification) to (trustee_feedback);

\end{tikzpicture}
\captionof{figure}{Trust Game payoffs.} \label{fig:trust_game_payoffs}
\end{center}
} 

Regarding the magnification factor, when $K > 1$, the value received back by trustor $u$ is positive and enables the necessary conditions for an extensive form of the trust game (long-term engagement). It becomes a strong indicator that trustee $v$ trustiness towards trustor $u$ is equal or above the minimum threshold $T$, where $\{T \in \mathbb{R} \mid 0 \leq T \leq 1 \}$.

When $0 \leq K < 1$, the service is causing the trustee some form of disruption (in the sense that efficiency has dropped below the level prior to using the service). This would be acceptable during the development phase of a product where the trustee takes part in a beta test program. In such a situation, the trustee sees a benefit in participating, assuming future value in adopting the service and the ability to harvest the benefits early on.

The worst-case scenario happens when $K < 0$. This could lead to rapid erosion of trustor $u$ trustworthiness, customer churn, and other negative outcomes.

\section{Quantifying Trust}
\label{section:quantifying_trust}

The aim of this trust game is to create the circumstances necessary for repeated interactions between trustor and trustee.

After the initial remittance $R_u$, given by \eqref{eq:trustor_initial_remittance}, there may be a residual value $r$ on the trustor's side that a trustee did not take advantage of. For instance, maybe not all product features are being used, inference happens in batches and data is yet to be sent through the pipeline, or some other reason. That residual value is what is left from $V$:

\begin{align} \label{eq:trustor_residual}
	r_u &= V - R_u \notag \\
	    &= V - p V \notag \\
	    &= (1 - p) V
\end{align}

The accumulated value $A$ for trustor $u$ upon completing the first cycle is the residual value $r_u$ \eqref{eq:trustor_residual} plus the repayment $B_u$ \eqref{eq:repayment_trustee} received from the trustee:

\begin{align} \label{eq:trustor_accumulated_value_1st_cycle}
	A_u^{\text{1st cycle}} &= r_u^1 + B_u^1 \notag \\
	    &= (1 - p_1) V + q_1 K_1 p_1 V \notag \\
	    &= V (1 - p_1 + q_1 K_1 p_1)
\end{align}

On the trustee's side, they will have received a value of $G_v$ \eqref{eq:trustee_initial_receival} and given back a portion $q$ of it. The net gain $N$ for trustee $v$ at the end of the first cycle is:

\begin{align} \label{eq:trustee_net_receival}
	N_v^{\text{1st cycle}} &= G_v^1 - q_1 G_v^1 \notag \\
	    &= (1 - q_1) K_1 p_1 V
\end{align}

Generalizing the gains for trustor and trustee for $n$ cycles of the trust game, we have equations for trustor:

\begin{align} \label{eq:trustor_accumulated_value_n_cycles}
	A_u &= V \left(1 - \sum_{i=1}^n p_i + \sum_{i=1}^n (q_i) \sum_{i=1}^n (K_i) \sum_{i=1}^n (p_i) \right)
\end{align}

and trustee:

\begin{align} \label{eq:trustee_net_receival_n_cycles}
	N_v &= V \left(1 - \sum_{i=1}^n q_i \right) \sum_{i=1}^n (K_i) \sum_{i=1}^n (p_i)
\end{align}

The objective is to maximize the payoff to the trustee and trustor---possibly skewed towards the trustee. As such, trust has to be repaid\cite{positive_political_economy} (\ie $q > 0$). The trustor benefits from economies of scale by the aggregate of payoffs from all trustees.

\section{Threshold}
\label{section:threshold}

For a trustor to increase its trustworthiness ($W_u$) in the eyes of a trustee, the gains delivered by the service must be higher than if the trustee was operating on their own. Such condition is satisfied by the following system of inequalities:

\begin{align} \label{eq:threshold_inequalities}
    W_u \subseteq 
    \begin{cases}
        p V \geq T \\
        K \geq 1
    \end{cases}
\end{align}

That happens when the value of the remittance $R_u$ is equal or greater than the threshold $T$ (the value sent is at a minimum equal to the perceived value received), and the magnification factor $K$ is greater or equal to one.

Being a system of inequalities, it is also possible to have a lower remittance ($p V < T$) and increase trustworthiness, as long as the magnification factor is large enough ($K \gg 1$) to make up for the shortfall. Although plausible, this would be uncommon.

\section{Simulations}
\label{section:simulated_experiments}

The following is a set of four simulations testing scenarios from fostering to eroding trust as a result of the quality of machine learning inference.

All the simulations begin from the same exact starting point, where it is assumed that the potential value of a product being offered to customers is of one million points (1,000,000). The starting number is an arbitrary value and could have been any positive number. We want to observe the shape of the curve formed from plotting interaction cycle after interaction cycle.

The hypothesis is that a trustee would increase their trustiness level towards the trustor by providing good machine learning inferences. Conversely, less than good enough results would have the opposite effect (\ie erode trust).

Notice that throughout all four simulations, all parameters are kept the same, varying only the magnification factor $K$.

\subsection{Simulation 1: Machine Learning Inferences Add Value}

\newcommand{\graphHeight}{118pt}
\newcommand{\graphWidth}{220pt}
\newcommand{\legendX}{0.67}
\newcommand{\legendY}{0.49}

For this simulation, we will go step-by-step in the first interaction. For subsequent simulations, only the final graph plots will be shown. Irrespective of the simulation, they all can be reproduced using the \href{https://gist.github.com/dcirne/8c74a2d8d5adaf59f9366a5212d41f22}{source code}\cite{jupyter-notebook} that accompanies this paper.


Assume that in the first cycle iteration, the trustor begins with $V = $ 1,000,000 points and is able to send a remittance of 65\% ($R_u = 0.65 \times$1,000,000) of inference value to a trustee. The magnification factor perceived by the trustee is $K = 2$, thus, the gain becomes 1,300,000 ($G_v = 2 \times$650,000) points.

The trustee sends a portion ($q = 0.14$) of the value back by interacting with the user interface, providing a feedback label, and paying for the service. The rebate received by the trustor is 182,000 ($B_u = 0.14 \times$1,300,000) points.

Adding the rebate to the residual value ($r_u = 0.35 \times$1,000,000), the trustor's accumulated gain equals 532,000 ($A_u =$ 350,000 $+$ 182,000) points, and the trustee's gain would be 1,118,000 ($N_v = 0.86 \times$ 1,300,000) points.

First, the trustee's perception was that they received more value than what the trustor had to offer due to the magnification factor (win). Second, the trustor received a rebate in various formats---accruing value that was not there before (win). And third, after the aggregate across all trustees, the trustor will have accumulated more than the initial value offered (win).

In Fig.~\ref{fig:accumulated_gains_K_>_1}, we can see the shape of the curve showing the accumulated gains for both trustor and trustee for the four cycles of the simulation.

{
\begin{center}
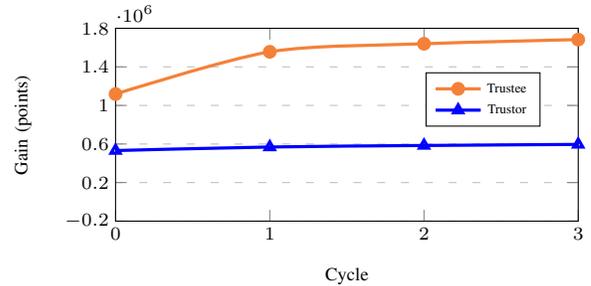

\begin{scriptsize}
\begin{tikzpicture}
\begin{axis}[
    xlabel={Cycle}, ylabel={Gain (points)},
    xmin=0, xmax=3, ymin=-200000, ymax=1800000,
    xtick={0, 1, 2, 3},
    ytick={-200000, 200000, 600000, 1000000, 1400000, 1800000},
    ymajorgrids=true, smooth,
    legend style={at={(\legendX, \legendY), font=\tiny}, anchor=south west},
    grid style=loosely dashed, height=\graphHeight, width=\graphWidth
]

\addplot[very thick, mark=*, color=TrusteeColor] plot
    coordinates {
    (0, 1118000.0)(1, 1556250.0)(2, 1639440.0)(3, 1683000.0)
    };

\addplot[very thick, mark=triangle, color=TrustorColor] plot
    coordinates {
    (0, 532000.0)(1, 568750.0)(2, 584560.0)(3, 595750.0)
    };

\legend{Trustee\\Trustor\\}

\end{axis}
\end{tikzpicture} 
\captionof{figure}{Accumulated gains $(K > 1)$.} \label{fig:accumulated_gains_K_>_1}
\end{scriptsize}
\end{center}
}

\subsection{Simulation 2: Machine Learning Inferences Are Neutral}

For the second simulation, a neutral magnification factor ($K = 1$) is being simulated. The value sent by the trustor and the value received by the trustee are perceived equally. The curve with the accumulated gains can be seen in Fig.~\ref{fig:accumulated_gains_K_=_1}. The trustee marginally sees an increase in the received value, whereas the trustor sees a small decline.

This scenario could be acceptable depending on the scale of the service and number of trustees, since the trustor's final gain is the aggregate from all trustees.

{
\begin{center}
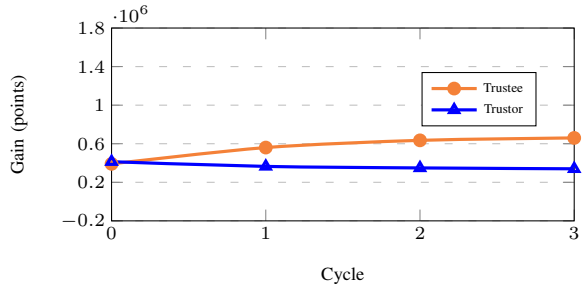

\begin{scriptsize}
\begin{tikzpicture}
\begin{axis}[
    xlabel={Cycle}, ylabel={Gain (points)},
    xmin=0, xmax=3, ymin=-200000, ymax=1800000,
    xtick={0, 1, 2, 3},
    ytick={-200000, 200000, 600000, 1000000, 1400000, 1800000},
    ymajorgrids=true, smooth,
    legend style={at={(\legendX, \legendY), font=\tiny}, anchor=south west},
    grid style=loosely dashed, height=\graphHeight, width=\graphWidth
]

\addplot[very thick, mark=*, color=TrusteeColor] plot
    coordinates {
    (0, 391300.0)(1, 560249.9)(2, 635040.0)(3, 660000.0)
    };

\addplot[very thick, mark=triangle, color=TrustorColor] plot
    coordinates {
    (0, 413699.9)(1, 364750.0)(2, 348959.9)(3, 339999.9)
    };

\legend{Trustee\\Trustor\\}

\end{axis}
\end{tikzpicture} 
\captionof{figure}{Accumulated Gains $(K = 1)$.} \label{fig:accumulated_gains_K_=_1}
\end{scriptsize}
\end{center}
}

\subsection{Simulation 3: Machine Learning Inferences Are Causing Inefficiencies}

The third simulation, Fig.~\ref{fig:accumulated_gains_0 <=_K_<_1}, shows a scenario where inefficiencies are being brought upon the trustee \mbox{($0 \leq K < 1$)}. Their gains are at best negligible, and at the same time there is a significant drop in the trustor's gains.

This situation would be plausible and acceptable only during the development phase of a product, where a trustee would have accepted to be an early adopter of the service.

{
\begin{center}
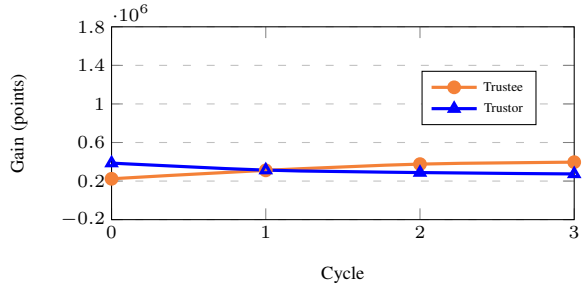

\begin{scriptsize}
\begin{tikzpicture}
\begin{axis}[
    xlabel={Cycle}, ylabel={Gain (points)},
    xmin=0, xmax=3, ymin=-200000, ymax=1800000,
    xtick={0, 1, 2, 3},
    ytick={-200000, 200000, 600000, 1000000, 1400000, 1800000},
    ymajorgrids=true, smooth,
    legend style={at={(\legendX, \legendY), font=\tiny}, anchor=south west},
    grid style=loosely dashed, height=\graphHeight, width=\graphWidth
]

\addplot[very thick, mark=*, color=TrusteeColor] plot
    coordinates {
    (0, 223600.0)(1, 311250.0)(2, 375840.0)(3, 396000.0)
    };

\addplot[very thick, mark=triangle, color=TrustorColor] plot
    coordinates {
    (0, 386399.9)(1, 313750.0)(2, 288160.0)(3, 273999.9)
    };

\legend{Trustee\\Trustor\\}

\end{axis}
\end{tikzpicture} 
\captionof{figure}{Accumulated Gains $(0 \leq K < 1)$.} \label{fig:accumulated_gains_0 <=_K_<_1}
\end{scriptsize}
\end{center}
}

\subsection{Simulation 4: Machine Learning Inferences Are Rapidly Eroding Trust}

The last simulation shows the worst-case scenario where machine learning inferences erode the trustor's trustworthiness ($K < 0$), reducing the trustee's ability to trust. Fig.~\ref{fig:accumulated_gains_K_<_0} shows how, in this scenario, there are negative gains (loss) for trustors and trustees. They are both worse off with the service, compared to operating without it.
    
{
\begin{center}
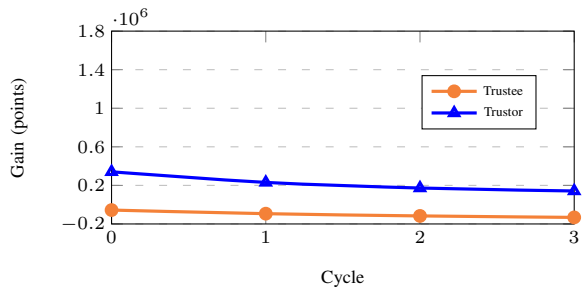

\begin{scriptsize}
\begin{tikzpicture}
\begin{axis}[
    xlabel={Cycle}, ylabel={Gain (points)},
    xmin=0, xmax=3, ymin=-200000, ymax=1800000,
    xtick={0, 1, 2, 3},
    ytick={-200000, 200000, 600000, 1000000, 1400000, 1800000},
    ymajorgrids=true, smooth,
    legend style={at={(\legendX, \legendY), font=\tiny}, anchor=south west},
    grid style=loosely dashed, height=\graphHeight, width=\graphWidth
]

\addplot[very thick, mark=*, color=TrusteeColor] plot
    coordinates {
    (0, -55900.0)(1, -93375.0)(2, -116640.0)(3, -132000.0)
    };

\addplot[very thick, mark=triangle, color=TrustorColor] plot
    coordinates {
    (0, 340900.0)(1, 230875.0)(2, 172639.9)(3, 141999.9)
    };

\legend{Trustee\\Trustor\\}

\end{axis}
\end{tikzpicture} 
\captionof{figure}{Accumulated Gains $(K < 0)$.} \label{fig:accumulated_gains_K_<_0}
\end{scriptsize}
\end{center}
}

\section{MEASURING AI AND ML RISK}
\label{section:measuring_ai_and_ml_risk}

One of the intended outcomes of quantifying trust is to define the metrics of risk. Then, it can be measured and monitored.

The National Institute of Standards and Technology (NIST) has published a study called \tqt{\href{https://nvlpubs.nist.gov/nistpubs/ai/NIST.AI.100-1.pdf}{Artificial Intelligence Risk Management Framework (AI RMF)}}\cite{nist-ai-rmf}. There, they claim that there is a finite set of traits that approximate to a good definition for a system to be trustworthy. We aim to extend the concepts to implement quantitative metrics and create a viable framework to monitor trustworthiness. NIST identifies seven broad categories. They are (The color-coded categories will be useful later in this paper when understanding an example of the framework implementation):

\begin{enumerate} \itemsep=-1pt
    \item {\color{CategoryColorReliability} Reliability and Validity}
    \item {\color{CategoryColorSafety} Safety}
    \item {\color{CategoryColorSecurity} Security and Resilience}
    \item {\color{CategoryColorAccountability} Accountability and Transparency}
    \item {\color{CategoryColorExplainability} Explainability and Interpretability}
    \item {\color{CategoryColorPrivacy} Privacy}
    \item {\color{CategoryColorFairness} Bias Management}
\end{enumerate}

For each of those categories, this paper proposes metrics that can be measured and used to compute a \emph{trust score}.

\subsection{Reliability and Validity}

A system is reliable when it does its job as intended, with minimal disruption of service\cite{ISO5723}, and when the results produced can be confirmed through objective evidence that the requirements were met\cite{ISO9001}. The following are proposed metrics for reliability and validity:

\begin{itemize} \itemsep=-1pt
    \item Uninterrupted uptime.
    \item Number of crashes.
    \item True Positives, True Negatives, False Positives, False Negatives.
    \item Latency between inquiry and returning results.
    \item Additionally, depending on the specific use case, the adoption of specific metrics (Accuracy, F1\cite{Lipton:2014aa}, BLEU\cite{bleu}, SuperGLUE\cite{superglue}, HELM\cite{helm}) is encouraged.
\end{itemize}

\subsection{Safety}

The state of the data, the system, the people, and the subject of inferences are not at a meaningful risk, that extends beyond physical safety. Those are metrics to represent that:

\begin{itemize} \itemsep=-1pt
    \item System design is represented in a diagram and is peer-reviewed, where appropriate.
    \item Data handling is done via a well-defined process with clear controls that align with existing regulations and oversight.
    \item A report that details to customers which data fields are used in training models.
    \item Access to the data is done with the consent of customers and is system-wide enforced by access roles.
    \item Once a model architecture is defined, models are trained using automation that does not require the intervention or participation of personnel.
\end{itemize}

\subsection{Security and Resilience}

Everyday operations have the ability to withstand adverse events or unexpected changes in the use or functioning of the environment.

\begin{itemize} \itemsep=-3pt
    \item Systems and people have explicit credentials to run and/or access the data.
    \item Isolation of data and systems from unauthorized agents.
    \item Implementation of multiple scopes of granted access/runtime, with each agent being assigned the minimum necessary level to perform a task.
\end{itemize}

\subsection{Accountability and Transparency}
\label{subsection:accountability_and_transparency}

Accountability and transparency of operations are necessary conditions for being trustworthy and increasing trust.

\begin{itemize} \itemsep=-3pt
    \item Report the data used in model training back to customers. The system must have a report, accessible by customers, that shows what data was used to train models.
\end{itemize}

\subsection{Explainable and Interpretable}

Explainability: the representation of the mechanisms underlying AI systems’ operations\cite{nist-ai-rmf}.

Interpretability: the meaning of AI systems’ outputs in the context of their designed functional purposes\cite{nist-ai-rmf}.

\begin{itemize} \itemsep=-3pt
    \item Identification of the principal component of inference results. 
    \item Displaying similar records can explain an inference by analogy.
    \item The ratio between the number of explanations given over the total number of explainable records would be the key metric.
\end{itemize}

\subsection{Privacy}

The norms and practices that help safeguard human autonomy, identity, and dignity. Freedom from intrusion, limiting observation, obtaining consent prior to disclosing or using Personally Identifiable Information (PII).

\begin{itemize} \itemsep=-2pt
    \item Definition and implementation of Legal, Privacy, and Responsibility frameworks.
    \item Transforming raw data into embeddings.
    \item De-identification and aggregation.
    \item Privacy awareness and training of the people involved.
\end{itemize}

\subsection{Bias Management}

Establish reasonable and viable frameworks for error prevention, then optimization of execution for error correction.

\begin{itemize} \itemsep=-3pt
    \item Number of reported and confirmed use cases.
    \item Subsequently, it can be offset by releasing new model versions that address those issues.
\end{itemize}

\section{IMPLEMENTATION}
\label{section:implementation}

Each of the metrics discussed in the \tqt{\nameref{section:measuring_ai_and_ml_risk}} section become numeric entries in a vector $M$, and associated with it, there is a stochastic vector $S$ containing weights representing how important each of the traits are in contributing to the creation of value and trustworthiness.

The dot product between $M$ and $S$ produces the \emph{Trust Score} $W$ \eqref{eq:measured_trust} which is our metric to signify value and trust.

\begin{align} \label{eq:measured_trust}
	W = M \cdot S^\mathsf{T}
\end{align}

The following is a simulated example of numeric scores attributed to items in each of the 7 categories. The names next to the entries of vector $M$ describe each item, as previously proposed, and the colors represent the categories.

The entries in vector $M$ related to True Positives (TP), True Negatives (TN), False Positives (FP), and False Negatives (FN) share Number of Inferences as common denominator. It is important to clarify that this is the total number of inferences where it is possible to categorize them as [TP, TN, FP, FN]. In many cases, the categorization of an inference is unknowable.

Note that the stochastic vector $S$ contains negative entries. Those are to penalize the corresponding metric in vector $M$ and reduce the trust score. For example, the higher the number of crashes, the lower the score.

{
\begin{scriptsize}
\begin{align} \label{eq:numeric_scores}
M = 
\begin{bNiceMatrix}[first-col]
    \text{\color{CategoryColorReliability}Uptime} & 99.99\% \\
    \text{\color{CategoryColorReliability}Number of Crashes} & 3 \\
    \text{\color{CategoryColorReliability}True Positives/Number of Inferences} & 60.00\% \\
    \text{\color{CategoryColorReliability}True Negatives/Number of Inferences} & 34.29\% \\
    \text{\color{CategoryColorReliability}False Positives/Number of Inferences} & 4.29\% \\
    \text{\color{CategoryColorReliability}False Negatives/Number of Inferences} & 1.43\% \\
    \text{\color{CategoryColorSafety}System Design} & 1 \\
    \text{\color{CategoryColorSafety}Data Handling Processes} & 1 \\
    \text{\color{CategoryColorSafety}Data Points Report} & 1 \\
    \text{\color{CategoryColorSafety}Data Access Consent} & 1 \\
    \text{\color{CategoryColorSafety}Touchless Model Training} & 1 \\
    \text{\color{CategoryColorSecurity}Access Control} & 1 \\
    \text{\color{CategoryColorSecurity}Tiered Access} & 1 \\
    \text{\color{CategoryColorSecurity}Data Isolation} & 1 \\
    \text{\color{CategoryColorAccountability}Data Usage Report} & 1 \\
    \text{\color{CategoryColorExplainability}Inference Explanation} & 40.00\% \\
    \text{\color{CategoryColorExplainability}Present Similar Records} & 20.00\% \\
    \text{\color{CategoryColorExplainability}Number of Explanation/Total Inferences} & 10.00\% \\
    \text{\color{CategoryColorPrivacy}Legal and Privacy Frameworks} & 1 \\
    \text{\color{CategoryColorPrivacy}De-identification of Data} & 0 \\    
    \text{\color{CategoryColorPrivacy}Privacy Training} & 1 \\
    \text{\color{CategoryColorFairness}Number of Confirmed Bias Issues} & 2 \\
    \text{\color{CategoryColorFairness}Number of Deployed Bias Fixes} & 1 \\
\end{bNiceMatrix}
\end{align}
\end{scriptsize}
}

\vspace{1pt}

{
\begin{scriptsize}
\begin{align} \label{eq:numeric_weights}
S = 
\begin{bmatrix}
    0.14 \\
    -0.14 \\
    0.24 \\
    0.24 \\
    -0.10 \\
    -0.10 \\
    0.01 \\
    0.02 \\
    0.02 \\
    0.01 \\
    0.04 \\
    0.06 \\
    0.07 \\
    0.06 \\
    0.05 \\
    0.05 \\
    0.05 \\
    0.05 \\
    0.06 \\
    0.05 \\
    0.02 \\
    -0.06 \\
    0.06 \\
\end{bmatrix}
\end{align}
\end{scriptsize}
}

Metrics of a qualitative nature are expressed numerically in vector $M$ as 0 or 1, representing their absence or presence. For instance, in the \tqt{\nameref{subsection:accountability_and_transparency}} section, we mention implementing a report that shows what data was used in training models. Either the report is available, or it is not. Although there may be degrees of completion of adoption in an organization, we are focused on the customer's perspective, that either the item is in place or it is not.

If possible, it will be good to keep the trust score within the [–1, 1] range, where –1 is the worst possible score, and 1 is the best score. We can use \eqref{eq:range_constraint} to apply this range constraint to the result of the \emph{trust score} computation.

\begin{align} \label{eq:range_constraint}
	W = min(1, max(W, -1))
\end{align}

In the case of the example provided in \eqref{eq:numeric_scores} and \eqref{eq:numeric_weights}, the trust score would be:

\begin{align} \label{eq:range_result}
	W = 0.635557 \notag
\end{align}

\subsection{Temporality}
\label{section:temporality}

The trust score $W$ is expected to display fluctuations over time. Since systems could experience an occasional malfunction, a model performance degradation, or an unanticipated incident, however, those fluctuations are presumed to be narrow and gentle, rather than wide and abrupt like a roller coaster.


It is plausible to imagine that after a few cycles of significant fluctuations in the \emph{trust score}, a customer would disengage and discontinue usage of the product.

\section{FAIR TRADING}
\label{section:fair_trading}

Fairness is an intrinsic concept associated with trust. Assuming that the trustor is providing value to a trustee, and in return the trustee is returning something of value to the trustor, the next step is to find that region of equilibrium where both parties accept the exchange as fair trade.

In addition, the region must be defined in such a way that it scales up or down proportionally to the exchange of value. For instance, imagine that a trustor went from providing one service, to providing two or three services; the trustor will expect to charge the trustee more. This section shows how this region of equilibrium is computed in such way that it remains a fair trade for both parties.

From \eqref{eq:trustor_accumulated_value_1st_cycle} and \eqref{eq:trustee_net_receival}, we know that the accumulated value $A_u$ by the trustor is the product’s residual value left, plus the repayment value sent by trustees. The net gain $N_v$ by trustees is the received magnified value, minus the repayment.

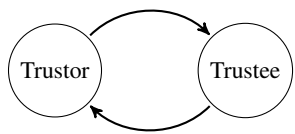
\begin{center}
\begin{tikzpicture}[-, >=stealth', shorten <= 1pt, shorten >= 1pt, auto, node distance=72pt, align=center]
                    
    \tikzstyle{every node}=[font=\small]

    \node (trustor) [draw, circle] {Trustor};
    \node (trustee) [draw, circle, right of=trustor] {Trustee};

    \path[->, in=135, out=45, thick] (trustor) edge (trustee);
    \path[->, in=315, out=225, thick] (trustee) edge (trustor);

\end{tikzpicture}
\label{fig:fair_trading}
\end{center}

\vspace{-15pt}

\begin{equation} \label{eq:next_state_accummulation}
A' = (1 - p)A + qN 
\end{equation}

\vspace{-18pt}

\begin{equation} \label{eq:next_state_net_gain}
N' = KpA - qN 
\end{equation}

From the previous paragraph, we see that \eqref{eq:next_state_accummulation} and \eqref{eq:next_state_net_gain} express how the next state of accumulation $A'$ and net gain $N'$ are computed. Expressing them in matrix format gives us \eqref{eq:next_state_matrix}.

\begin{equation} \label{eq:next_state_matrix}
\begin{pmatrix}
    A' \\
    N'
\end{pmatrix}
= 
\begin{pmatrix}
    1-p & q \\
    Kp & -q
\end{pmatrix}
\begin{pmatrix}
    A \\
    N
\end{pmatrix}
\end{equation}

We want to find that region of values that would make the trade between trustor and trustee to be considered fair.

From Linear Algebra, we know that the eigenvectors\cite{linear-algebra} of a matrix will give us the space that could scale---but otherwise would remain unchanged---irrespective of the linear transformation applied to it (assuming that the eigenvectors are linearly independent and have no imaginary $i$ component).

Given that the conditions are satisfied, the linear transformation would be the addition or subtraction of services and the proportional increase or decrease of charges and feedback interactions, in other words, scaling up, down, or neutral.

The eigenvector associated with the largest, positive eigenvalue of the matrix shown in \eqref{eq:next_state_matrix} can be interpreted as the region where both parties should consider transactions between them as fair trade, thus contributing to preventing the erosion of trust.

Let us build an example. Assume that the percentage of remitted value $p$, the repayment portion $q$, and magnification factor $K$ have the following values:

\begin{align} \label{eq:example_values}
&p = 0.85, \, q = 0.14, \, K = 2 \notag
\end{align}

Substituting these values in the matrix from \eqref{eq:next_state_matrix} leads us to:

\begin{equation} \label{eq:next_state_matrix_values}
\begin{pmatrix}
    A' \\
    N'
\end{pmatrix}
= 
\begin{pmatrix}
    0.15 & 0.14 \\
    1.7 & -0.14
\end{pmatrix}
\begin{pmatrix}
    A \\
    N
\end{pmatrix}
\end{equation}

One condition that needs to be satisfied is that the vectors---derived from the matrix in \eqref{eq:next_state_matrix_values}---are linearly independent, so they can span the space being considered. Otherwise, they would only represent a sub-space and not necessarily produce the fair trade region we aim for.

As you can see in Fig.~\ref{fig:linearly_independent}, the vectors are linearly independent and also satisfy the other conditions to compute the eigenvectors to determine the fair trade region between trustor and trustee.

\begin{center}
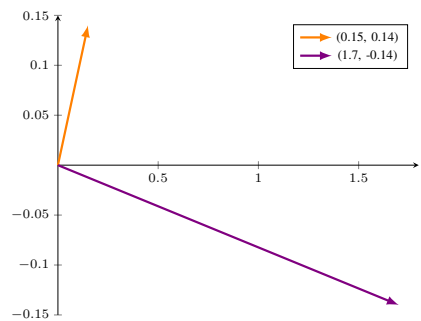

\begin{scriptsize}
\begin{tikzpicture}[scale=0.70]
\begin{axis}[
    axis lines=center,
    xmin = 0, xmax = 1.8,
    ymin = -0.15, ymax = 0.15,
    xtick distance = 0.5,
    ytick distance = 0.05,
    legend pos=north east,
    scaled ticks=false, tick label style={/pgf/number format/fixed}
]
 
\addplot[->, very thick,  orange, -latex] coordinates {(0, 0) (0.15, 0.14)};

\addplot[->, very thick,  violet, -latex] coordinates {(0, 0) (1.7, -0.14)};

\legend{(0.15, 0.14)\\(1.7, -0.14)\\}

\end{axis}
\end{tikzpicture}
\captionof{figure}{Linearly independent vectors.} \label{fig:linearly_independent}
\end{scriptsize}
\end{center}

The next step is to compute its eigenvalues and eigenvectors, then find the line defined by the eigenvector associated with the largest eigenvalue.

\begin{align}
\lambda^{\downarrow}_{1} &= 0.513945 \label{eq:eigenvalue1} \\
\lambda_{2} &= -0.503945 \label{eq:eigenvalue2} \\
E_{\lambda_1} &= \begin{pmatrix}
    0.384674 \\
    1
\end{pmatrix} \label{eq:eigenvector1} \\
E_{\lambda_2} &= \begin{pmatrix}
    -0.214085 \\
    1
\end{pmatrix} \label{eq:eigenvector2}
\end{align}

The largest eigenvalue is $\lambda^{\downarrow}_{1}$, thus our eigenvector of interest is $E_{\lambda_1}$. In order to find the line defined by $E_{\lambda_1}$ coordinates, we just need to compute its slope, since the eigenvector starts at the origin (0, 0).

\begin{align}
y &= mx + b \label{eq:slope} \\
m &= \cfrac{1 - 0}{0.384674 - 0} = 2.599604 \label{eq:slope_m} \\
b &= 0 \label{eq:slope_shift} \\
y &= 2.599604 \, x \label{eq:slope_result}
\end{align}

Fig.~\ref{fig:fair_trade_region} shows the eigenvector $E_{\lambda_1}$ in {\color{red} red} and the line derived by it, and defined by \eqref{eq:slope_result}, in {\color{blue} blue}. The line characterizes the fair trade region since any point on it carries the maximum accumulated value $A$ and net gains $N$, for the trustor and trustee, respectively.

\begin{center}
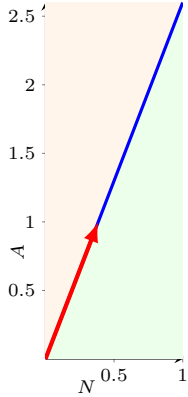

\begin{scriptsize}
\begin{tikzpicture}[scale=0.60]
    \begin{axis}[
        axis lines=center,
        axis equal image,
        xtick distance = 0.5,
        ytick distance = 0.5,
        width=110mm,
        view = {0}{90},
        no marks,
        xmin = 0, xmax = \domainX,
        ymin = 0, ymax = \domainY,
        xlabel = {$N$},
        ylabel = {$A$},
        x label style={at={(axis description cs:0.5, -0.07)}, anchor=north},
        y label style={at={(axis description cs:-0.18, 0.5)}, rotate=90, anchor=south},
    ]
    
    \addplot[draw=none, name path=ceiling, domain = 0:\domainX] {\domainY};
    \addplot[draw=none, name path=x_axis, domain = 0:\domainX] {0};
    \addplot[orange!8] fill between[of=ceiling and x_axis, soft clip={domain=0:\domainX}];
    \addplot[smooth, blue, very thick, name path=trust_line, domain = 0:\domainX] {2.599604 * x};
    \addplot[green!8] fill between[of=trust_line and x_axis, soft clip={domain=0:\domainX}];
    \addplot[->, ultra thick,  red, -latex] coordinates {(0, 0) (0.384674, 1)};

    \end{axis}
\end{tikzpicture}
\captionof{figure}{Fair trade region.} \label{fig:fair_trade_region}
\end{scriptsize}
\end{center}

The colored areas above and below the line represent the regions where either the trustor would accumulate more value ({\color{orange!70} orange}) or the trustee would retain more gains ({\color{Green!80} green}).

\section{Conclusion}
\label{section:conclusion}

This paper takes a step forward in contributing to the conversation about trust in ML-based systems. It presented a realistic and viable framework to compute a trust score and demonstrated that good machine learning inference results satisfy a valid criterion to increase a trustor’s trustworthiness, allowing for trustees to be more trusting.

A strong motivation exists to provide inferences only when a minimum confidence level has been cleared. It would be preferable to not produce a result than to provide a low-confidence one. When nothing is provided, a customer can still operate at their nominal level of productivity.

We established the items of interest for measuring, defined a system to compute and weigh each contribution, and identified the region of fair trade where win-win relationships between trustor and trustee can take place and scale up or down.

Trust has a temporal nature to it; its behavior is not linear, but instead it is expected to oscillate with gentle fluctuations. Trust and value add are not only earned, but also require maintenance over time.

Lastly, we demonstrated that it is possible to establish a region of fair trading where both trustors and trustees perceive fairness in the exchange of value.


\end{document}